\documentclass[10pt,twocolumn,letterpaper]{article}

\usepackage[pagenumbers]{cvpr} 

\definecolor{cvprblue}{rgb}{0.21,0.49,0.74}
\usepackage[pagebackref,breaklinks,colorlinks,allcolors=cvprblue]{hyperref}
\usepackage{multirow}
\usepackage{makecell}
\usepackage{bbding}
\usepackage{pifont}
\usepackage[accsupp]{axessibility}
\def\paperID{41453} 
\def\confName{CVPR}
\def\confYear{2026}

\title{D2Dewarp: Dual Dimensions Geometric Representation Learning Based \\ Document Image Dewarping}
\author{
Heng Li\textsuperscript{1},
Xiangping Wu\textsuperscript{1}\footnotemark[1],
Qingcai Chen\textsuperscript{1, 2}\footnotemark[1]
\and
\textsuperscript{1}Harbin Institute of Technology Shenzhen, China, 
\textsuperscript{2}PengCheng Laboratory, Shenzhen China
\and
{\tt\small hengli.lh@outlook.com, wxpleduole@gmail.com, qingcai.chen@hit.edu.cn}}

\begin{document}
\maketitle
\renewcommand{\thefootnote}{\fnsymbol{footnote}}
\footnotetext[1]{Corresponding authors.}
\renewcommand{\thefootnote}{\arabic{footnote}}


\begin{abstract}
Document image dewarping remains a challenging task in the deep learning era. While existing methods have improved by leveraging text line awareness, they typically focus only on a single horizontal dimension. In this paper, we propose a fine-grained deformation perception model that focuses on \textbf{D}ual \textbf{D}imensions of document horizontal-vertical-lines to improve document \textbf{Dewarp}ing called \textbf{D2Dewarp}. It can perceive distortion trends in different directions across document details. To combine the horizontal and vertical granularity features, an effective fusion module based on X and Y coordinate is designed to facilitate interaction and constraint between the two dimensions for feature complementarity. Due to the lack of annotated line features in current public dewarping datasets, we also propose an automatic fine-grained annotation method using public document texture images and automatic rendering engine to build a new large-scale distortion training dataset named \textbf{DocDewarpHV}. On three public Chinese and English benchmarks, both quantitative and qualitative results show that our method achieves better rectification results compared with the state-of-the-art methods. The code and dataset are available at \url{https://github.com/xiaomore/D2Dewarp}. 
\end{abstract}

\section{Introduction}
In recent years, the widespread usage of mobile electronic devices, such as smartphones and cameras, has led people to rely increasingly on capturing electronic documents through photography. However, issues such as device placement, lighting conditions, and paper deformation often result in varying degrees of distortion in the captured document images. As a result, extracting useful information from these images becomes challenging, posing difficulties for tasks such as degraded image restoration \cite{Zhang2024DocResAG, tie2025local, yang2023docdiff}, text content recognition \cite{Li2023CharacterAwareSA, Peng2022RecognitionOH, Wei2024GeneralOT}, document understanding \cite{Fang2020ACS, Jin2020RUArtAN, Feng2023DocPediaUT, Liao2024DocLayLLMAE, Zhang2024DocKylinAL}. Consequently, the rectification of distorted document images has garnered increasing attention from researchers. In this work, we focus on dewarping arbitrarily deformed document images captured in real-world scenarios. 

\begin{figure}[t]
  \centering
  \includegraphics[width=0.95\linewidth]{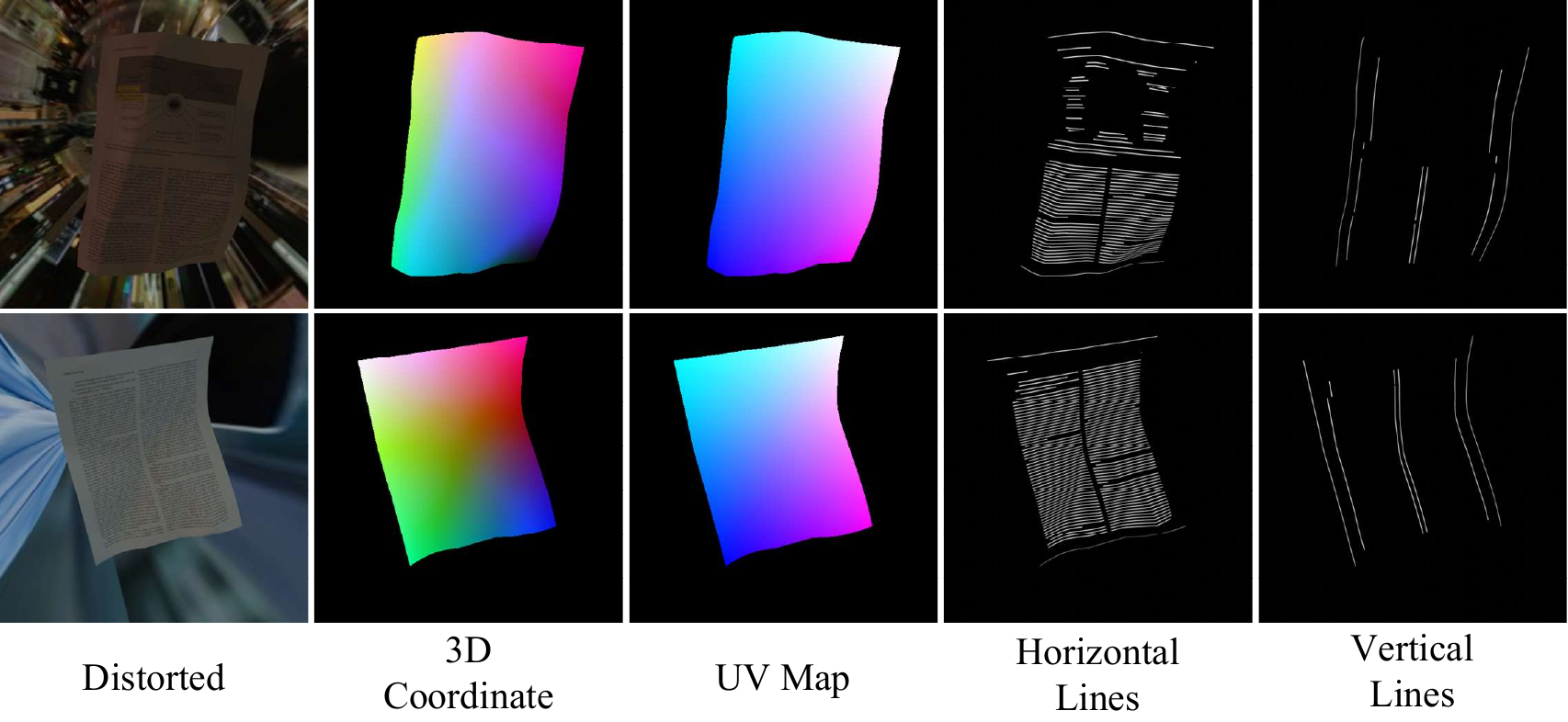}
  \caption{Visualization of our proposed dataset \emph{DocDewarpHV}. The first column is the distorted document image with complex background. \emph{3D Coordinate}: the position of each pixel of the distorted image in three-dimensional space. \emph{UV map}: maps the 3D surface to the location of the 2D coordinate system texture map. \emph{Horizontal Lines}: in addition to text lines, its also include the top and bottom boundaries of documents, tables, figures, and paragraphs. \emph{Vertical Lines}: the left and right boundaries of the deformed document area, tables, figures, and paragraphs.}
  \label{fig:3views}
\end{figure}

Recently, representation learning-based rectification techniques have significantly enhanced the visualization and readability of distorted document images. 
These methods employed convolutional neural networks (CNNs) and transformers to learn 2D deformation field  (backward map) that maps the warped image to the scanned image, thereby optimizing the geometric dewarping task \cite{Ma2018DocUNetDI, Das2019DewarpNetSD, Liu2020GeometricRO, Xie2020DewarpingDI, Feng2021DocTrDI, Xie2022DocumentDW, Jiang2022RevisitingDI, Feng2022GeometricRL, Dai2023MataDocMA, Li2023LayoutawareSD, Li2023ForegroundAT, Yu_2024_WACV, 10327775, hertlein2025docmatcher, zhang2025dvd, zhang2025enhancing, han2025docmamba, Zhang2022MariorMR}. For example, DocUNet \cite{Ma2018DocUNetDI} utilized UNet networks to regress the 2D backward map. 
DDCP \cite{Xie2022DocumentDW} developed a method to predict the mapping between control points and reference points, convert the sparse mapping into a backward mapping through interpolation, and remap the original distorted document image to a corrected version. 
DocReal \cite{Yu_2024_WACV} combined channel and spatial attention to enhance control point prediction to better capture local deformations.
These methods utilized CNNs or attention mechanism to directly regress 2D deformation fields but cannot effectively integrate the multi-granular features of the document, lacking the information guidance of the internal attributes.

To alleviate the above issues, some methods incorporated additional information to assist in dewarping. On the one hand, global information is introduced as an auxiliary. DewarpNet \cite{Das2019DewarpNetSD}, DocGeoNet \cite{Feng2022GeometricRL} and UVDoc \cite{UVDoc} utilized an end-to-end networks to predict 3D coordinates and 2D deformation fields. DocTr \cite{Feng2021DocTrDI}, Marior \cite{Zhang2022MariorMR}, DocGeoNet\cite{Feng2022GeometricRL}, FCN-based \cite{Xie2020DewarpingDI} and FTDR \cite{Li2023ForegroundAT} mitigated complex background interference through pixel-level foreground extraction network.
More fine-grained, on the other hand, local information is introduced as an auxiliary, recent approaches enhanced document readability by focusing on text line regions such as RDGR \cite{Jiang2022RevisitingDI}, DocGeoNet \cite{Feng2022GeometricRL}, FTDR \cite{Li2023ForegroundAT} and layout information like LA-DocFlatten \cite{Li2023LayoutawareSD} to further refine local rectification in distorted images. 

These excellent methods demonstrate that text lines and layout information improve document rectification. However, their granularity lacks mutual constraints and complementarity. Text lines based approaches overlook tabular and graphical elements, while layout-guided methods treat figures, tables, and paragraphs as isolated categories, neglecting finer intra-category interactions such as paragraph-level text-line relationships.

To address these limitations, we define image deformation representation through horizontal and vertical dimensions. For distorted document images containing layout elements (illustrated in Figure~\ref{fig:3views}), we classify the top and bottom boundaries of the following as horizontal lines: the entire document foreground, tables, figures, paragraphs, and their constituent text lines. Similarly, left and right boundaries form vertical lines. Based on bidirectional deformation characteristics, we propose a fusion module to effectively learn complementary constraints between these dimensional features.

The main contributions of our work are summarized as follows:

\begin{itemize}
\item We present a novel end-to-end architecture for dual-dimension geometric representation learning of warped document images, capturing fine-grained deformation trends in both horizontal and vertical directions.

\item We design an effective fusion module to integrate distortion features from both dimensions, enhancing their interaction and constraints for optimal feature complementarity.

\item We release a new training dataset of distorted document images with more refined annotations in dual dimensions. Experimental results confirm that our method attains state-of-the-art performance. We will publicly release our model and source code to facilitate further research and development.
\end{itemize}

\section{Related Work}

\textbf{Traditional Methods.} Document image rectification serves not only for visual enhancement but also facilitates downstream tasks including document understanding and text extraction. Traditional approaches typically employ parametric regression based on superficial features like text lines \cite{Huang2015TextLE}, cylindrical surfaces \cite{Wada1997ShapeFS, Courteille2007ShapeFS, Liang2008GeometricRO, Meng2018ExploitingVF}, and document boundaries \cite{Tsoi2007MultiViewDR, Brown2006GeometricAS}.

\textbf{Deep Learning Methods.} Recent advances leverage deep representation learning and novel architectures. DocUNet \cite{Ma2018DocUNetDI} regresses 2D deformation fields using synthetic data and stacked UNets. DocTr \cite{Feng2021DocTrDI} pioneers transformer-based dewarping with foreground extraction. DocScanner \cite{feng2025docscanner} implements recurrent stepwise rectification. DocTr-plus \cite{feng2023doctrp} enhances DocTr through hierarchical multi-scale parsing. DRNet \cite{10327775} introduces coarse-to-fine processing with consistency loss.

\textbf{Global or Local Information Utilization Methods.} DewarpNet \cite{Das2019DewarpNetSD} models global 3D coordinates via UNet. Piece-wise \cite{Das2021EndtoendPU} fuses local deformation fields with global coordinates. DocGeoNet \cite{Feng2022GeometricRL} and RDGR \cite{Jiang2022RevisitingDI} integrate text line attributes. FTDR \cite{Li2023ForegroundAT} extends these via foreground-textline cross-attention. LA-DocFlatten \cite{Li2023LayoutawareSD} incorporates a hybrid segmentation network, utilizing both convolutional neural networks and transformers to extract layout information from documents. And a regression module designed to predict UV maps.

Unlike existing approaches, our model emphasizes simultaneous horizontal-vertical feature interaction. Our framework achieves fine-grained fusion through two dimensions constraint learning, focusing on mutual feature complementarity beyond isolated global or local optimization.

\begin{figure*}
\begin{center}
\includegraphics[width=0.9\linewidth]{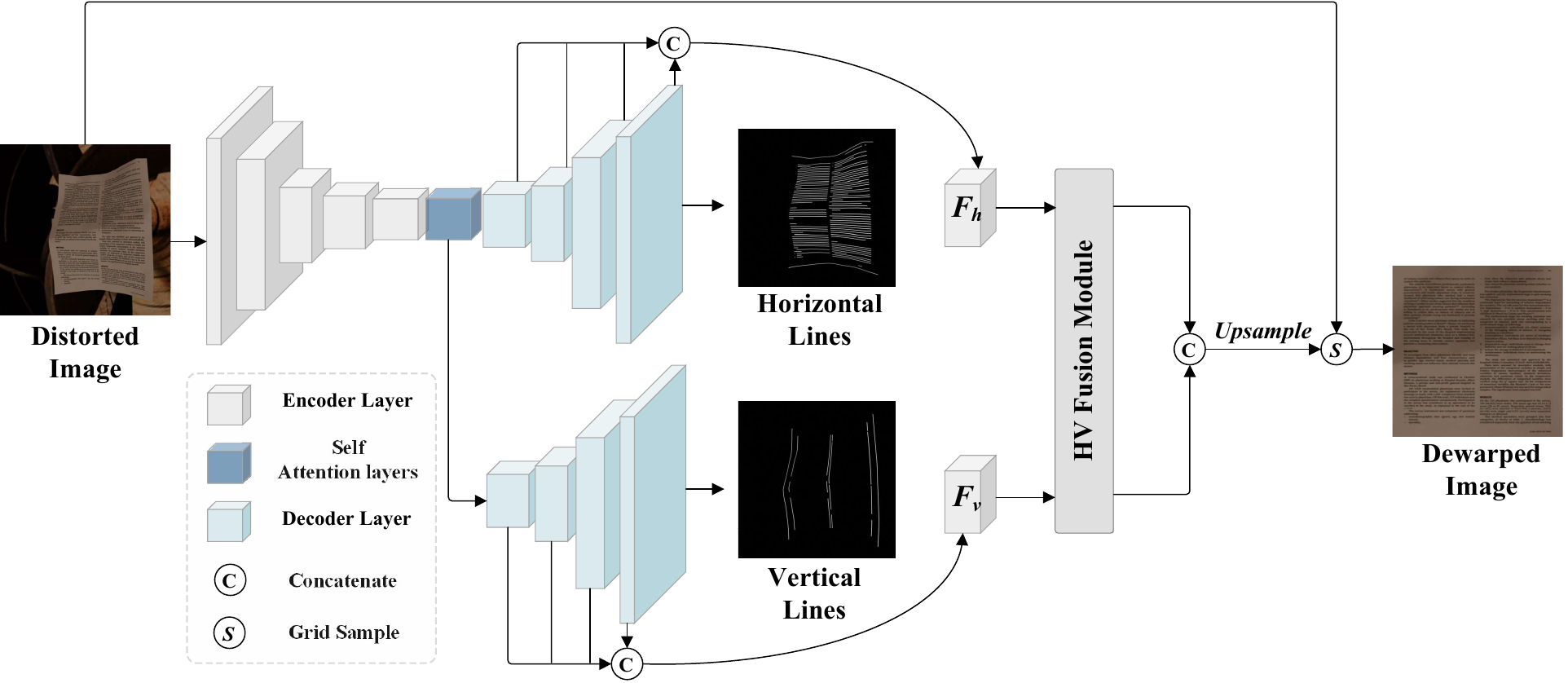}
\end{center}
  \caption{The architecture of our proposed method {\itshape D2Dewarp}. The segmentation model of the UNet structure predicts the two dimensions of lines. The dual decoders share the same encoder. Each layer in the decoder outputs the prediction result of the line, and then the feature map of each layer is resized to one-eighth of the input image and concatenated to obtain {\itshape $F_h$} and {\itshape $F_v$} respectively. The HV Fusion Module is used to fuse the feature maps of horizontal and vertical lines. For better visualization, we omit the skip connection line of UNet in this figure.}
  \vspace{-2mm}
  \label{fig:model_overview}
\end{figure*}

\section{Proposed Method}
In this section, we introduce the overall structure of our proposed method for distorted document image dewarping. The architecture of the model is shown in Figure~\ref{fig:model_overview}. Our network mainly consists of two parts: the horizontal and vertical line segmentation model of the UNet structure with dual decoders, and a lightweight module (HV Fusion Module) that fuses these two line features.

\textbf{Segment Dual Lines}. Drawing on the excellent effect of UNet \cite{ronneberger2015u} in the field of image segmentation, we use a similar encoder structure in this part. For an input distorted document image $\boldsymbol{I} \in \mathbb {R} ^ {H\times W\times C}$, $H$=$W$=$448$ is the height and width of the image, and $C$=$3$ denotes the RGB channels of the source distorted image. It is input into the segmentation model to predict horizontal and vertical lines. The encoder contains five layers. 
The input image undergoes two $3 \times 3$ convolutions, batch normalization (BN) \cite{Ioffe2015BatchNA}, and Rectified Linear Unit (ReLU) activation \cite{Nair2010RectifiedLU} (ConvBNReLU) in the first layer to extract general feature representations. The remaining four layers are downsampling modules, which consist of maximum pooling and two ConvBNReLU. 
The feature map output size of the last layer of the encoder is $1/16$ of the input image $\boldsymbol{I}$. The feature map output by the encoder are recorded as \{$E_1, E_2, E_3, E_4, E_5$\} with the output size \{$448\times448, 224\times224, 112\times112, 56\times56, 28\times28$\} and the channels \{$32, 64, 128, 196, 448$\}. In order to capture the long-distance dependency between feature map elements, we use $l$ layers of self-attention after the encoder, where $l=4$.

\textbf{Dual Decoders}. Each decoder has four layers with exactly the same structure, upsampling and line prediction modules. Among them, the upsampling module consists of bilinear interpolation and two ConvBNReLU. Line prediction is implemented by a layer of $1\times1$ convolution and then directly upsample to the input image size using bilinear interpolation. The output feature map of the decoder layer are recorded as \{$D_1, D_2, D_3, D_4$\} with the output size \{$56\times56, 112\times112, 224\times224, 448\times448$\} and the channels \{$196, 128, 64, 48$\}. These four feature maps are resized to one eighth of the input image, then concatenated in the channel dimension and input into ConvBNReLU to obtain horizontal and vertical geometric features $\boldsymbol{F}_h \in \mathbb {R} ^ {56\times 56\times 448}$ and $\boldsymbol{F}_v \in \mathbb {R} ^ {56\times 56\times 448}$ respectively.

\begin{figure}[t]
\begin{center}
\includegraphics[width=0.95\linewidth]{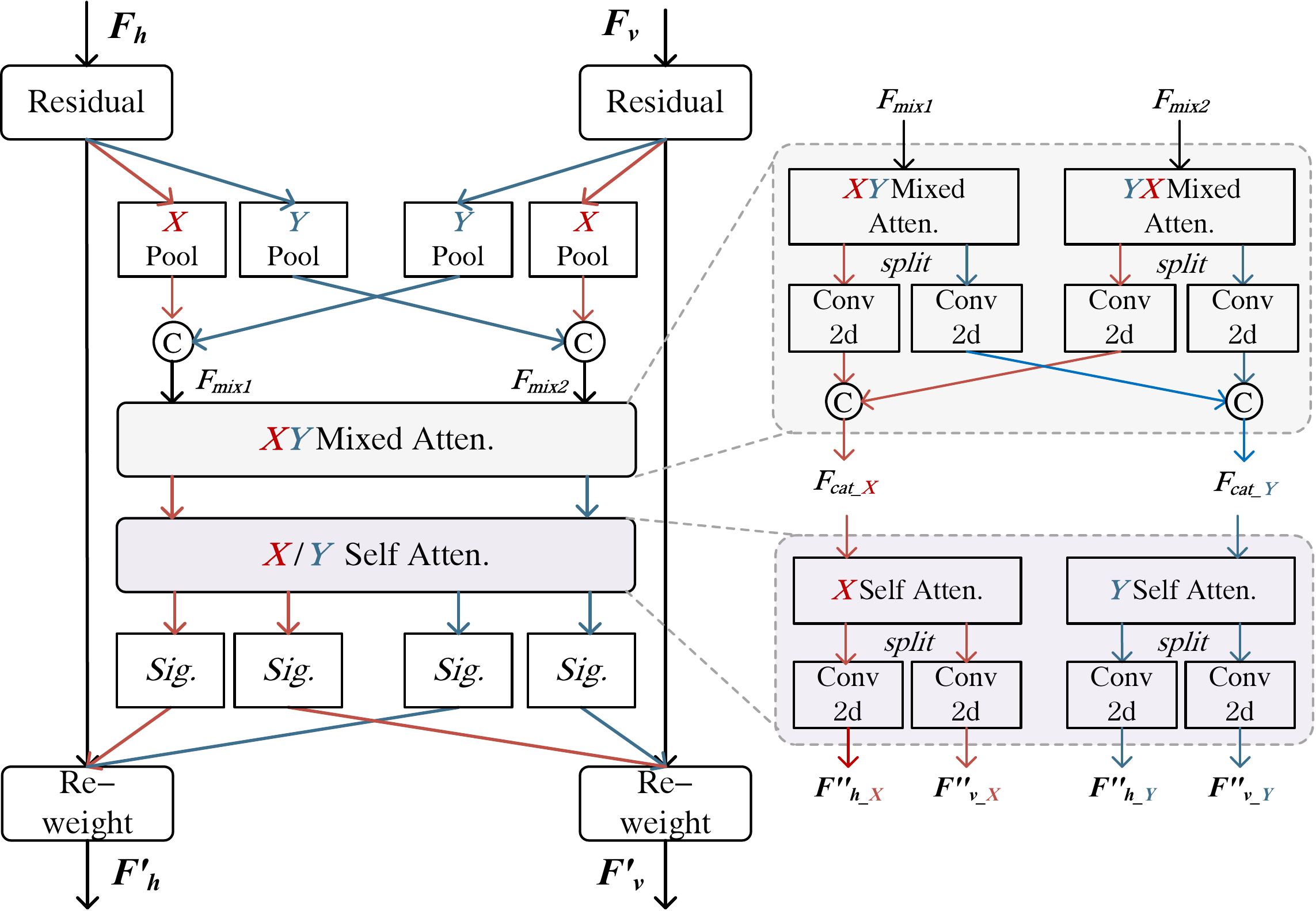}
\end{center}
  \caption{The HV Fusion Module. $\boldsymbol{F}_h$ and $\boldsymbol{F}_v$ represent the horizontal and vertical feature maps obtained by the segmentation model as the input of this module. \textcolor{purple} {\textbf{\textit{X}}} and \textcolor{cyan} {\textbf{\textit{Y}}} Pool are $AvgPool$ in Equation (\ref{equal:avgpool}) using $AdaptiveAvgPool2d$. $Sig.$ refers to Sigmoid activation function. $C$ denotes concatenation. Arrows in \textcolor{purple} {red} and \textcolor{cyan} {blue} indicate the pathways of feature fusion in the \textcolor{purple} {\textbf{\textit{X}}} and \textcolor{cyan} {\textbf{\textit{Y}}} directions.} 
  \label{fig:fusion_module}
\end{figure}

\textbf{HV Fusion Module}. This module aims to fuse the two geometric features obtained by the segmentation model. Leveraging the pooling operation's ability to aggregate local information on the feature map and inspired by the superiority of coordinate attention in image classification \cite{Hou_2021_CVPR}, we propose using 2D average pooling to capture local information along the $X$ and $Y$ spatial directions. For feature maps, the so-called $X$ and $Y$ also refer to the two dimensions of width and height. The horizontal $\boldsymbol{F}_h$ and vertical features $\boldsymbol{F}_v$ of the distorted image obtained by the segmentation model correspond to the pooling in the $X$ and $Y$ directions. The module is shown in Figure~\ref{fig:fusion_module}. For $\boldsymbol{F}_h$ and $\boldsymbol{F}_v$, $X$ and $Y$ two-dimensional average pooling is used to obtain the mixed pooling features:

\begin{equation}
\begin{aligned}
\label{equal:avgpool}
  &\boldsymbol{F}_{h\_\textcolor{purple}{\textbf{\textit{X}}}}, \boldsymbol{F}_{h\_{\textcolor{cyan}{\textbf{\textit{Y}}}}}={AvgPool_{\textcolor{purple}{\textbf{\textit{X}}}}}(\boldsymbol{F}_h), {AvgPool_{\textcolor{cyan}{\textbf{\textit{Y}}}}}(\boldsymbol{F}_h) \\
  &\boldsymbol{F}_{v\_{\textcolor{cyan}{\textbf{\textit{Y}}}}}, \boldsymbol{F}_{v\_{\textcolor{purple}{\textbf{\textit{X}}}}}={AvgPool_{\textcolor{cyan}{\textbf{\textit{Y}}}}}(\boldsymbol{F}_v), {AvgPool_{\textcolor{purple}{\textbf{\textit{X}}}}}(\boldsymbol{F}_v) \\
  &\boldsymbol{F}_{{mix1}}=CAT(\boldsymbol{F}_{h\_{\textcolor{purple}{\textbf{\textit{X}}}}}, Trans(\boldsymbol{F}_{v\_{\textcolor{cyan}{\textbf{\textit{Y}}}}}))_{dim=2} \\
  &\boldsymbol{F}_{{mix2}}=CAT(Trans(\boldsymbol{F}_{h\_{\textcolor{cyan}{\textbf{\textit{Y}}}}}), \boldsymbol{F}_{v\_{\textcolor{purple}{\textbf{\textit{X}}}}})_{dim=2} \\
\end{aligned}
\end{equation}
\noindent where $CAT(\cdot)$ represents feature map concatenation. $Trans(\cdot)$ is dimension transposition. After $AvgPool$, the width size of $\boldsymbol{F}_{h\_\textcolor{purple}{\textbf{\textit{X}}}}$ and the height of $\boldsymbol{F}_{h\_{\textcolor{cyan}{\textbf{\textit{Y}}}}}$ is set to 1.

The mixed attention mechanism is used to capture the long-distance dependencies in the $X$ or $Y$ direction from $\boldsymbol{F}_h$ or $\boldsymbol{F}_v$ respectively. Interacting features from different sources in different directions can achieve the purpose of constraining horizontal and vertical features to each other, thereby making up for their respective shortcomings. The mixed attention ($MixedAtten$ in Figure~\ref{fig:fusion_module}) here refers to the attention relationship between features in different directions ($X$ and $Y$) of the two feature maps. The calculation formulas are as follows:
\begin{equation}
\begin{aligned}
&\boldsymbol{F}_{{mix1}}^{\prime}=MixedAtten(\boldsymbol{F}_{{mix1}}) \\
&\boldsymbol{F}_{{mix2}}^{\prime}=MixedAtten(\boldsymbol{F}_{{mix2}}) \\
&\boldsymbol{F}_{h\_{\textcolor{purple}{\textbf{\textit{X}}}}}^{\prime}, \boldsymbol{F}_{v\_{\textcolor{cyan}{\textbf{\textit{Y}}}}}^{\prime}=Split(\boldsymbol{F}_{{mix1}}^{\prime}) \\
&\boldsymbol{F}_{v\_{\textcolor{purple}{\textbf{\textit{X}}}}}^{\prime}, \boldsymbol{F}_{h\_{\textcolor{cyan}{\textbf{\textit{Y}}}}}^{\prime}=Split(\boldsymbol{F}_{{mix2}}^{\prime}) \\
&\boldsymbol{F}_{cat\_{\textcolor{purple}{\textbf{\textit{X}}}}}=CAT(Conv2d(\boldsymbol{F}_{h\_{\textcolor{purple}{\textbf{\textit{X}}}}}^{\prime}), Conv2d(\boldsymbol{F}_{v\_{\textcolor{purple}{\textbf{\textit{X}}}}}^{\prime})) \\
&\boldsymbol{F}_{cat\_{\textcolor{cyan}{\textbf{\textit{Y}}}}}=CAT(Conv2d(\boldsymbol{F}_{h\_{\textcolor{cyan}{\textbf{\textit{Y}}}}}^{\prime}), Conv2d(\boldsymbol{F}_{v\_{\textcolor{cyan}{\textbf{\textit{Y}}}}}^{\prime})) \\
\end{aligned}
\end{equation}
\noindent where $Split(\cdot)$ represents channel separation.
After obtaining the mixed attention features $\boldsymbol{F}_{{mix1}}^{\prime}$ and $\boldsymbol{F}_{{mix2}}^{\prime}$, let the features return to different directions of independent line features. In this step, we separate them in the channel dimension. $X/Y$ self attention is used to capture the long-distance dependencies in the $X$ and $Y$ directions from $\boldsymbol{F}_h$. The same applies to $\boldsymbol{F}_v$.
$X/Y$ self attention here refers to the attention relationship between the features of two line feature maps in the same direction ($X$ or $Y$). Formulas are as follows,
 where $Sig(\cdot)$ represents the $Sigmoid$ activation function, which normalizes feature values to $[0, 1]$. Concatenate $\boldsymbol{F}_h^{\prime}$ and $\boldsymbol{F}_v^{\prime}$ in the channel dimension, and then predict the 2D deformation field $\boldsymbol{\hat{G}}$ through the upsampling module. Here we use the upsampling module of DocTr, DocGeoNet and FTDR. The 2D deformation field has two channels, mapping the vertical and horizontal position coordinates between the distorted image pixels and the flat image.
\begin{equation}
\begin{aligned}
  &\boldsymbol{F}_{h\_{\textcolor{purple}{\textbf{\textit{X}}}}}^{\prime\prime}, \boldsymbol{F}_{v\_{\textcolor{purple}{\textbf{\textit{X}}}}}^{\prime\prime}= Split({\textcolor{purple}{\bf \emph{X}}}SelfAtten(\boldsymbol{F}_{cat\_{\textcolor{purple}{\textbf{\textit{X}}}}})) \\
  &\boldsymbol{F}_{h\_{\textcolor{cyan}{\textbf{\textit{Y}}}}}^{\prime\prime}, \boldsymbol{F}_{v\_{\textcolor{cyan}{\textbf{\textit{Y}}}}}^{\prime\prime}=Split({\textcolor{cyan}{\bf \emph{Y}}}SelfAtten(\boldsymbol{F}_{cat\_{\textcolor{cyan}{\textbf{\textit{Y}}}}})) \\
  &\boldsymbol{F}_{h\_{\textcolor{purple}{\textbf{\textit{X}}}}}^{\prime\prime}, \boldsymbol{F}_{v\_{\textcolor{purple}{\textbf{\textit{X}}}}}^{\prime\prime}=Sig(Conv2d(\boldsymbol{F}_{h\_{\textcolor{purple}{\textbf{\textit{X}}}}}^{\prime\prime})), Sig(Conv2d(\boldsymbol{F}_{v\_{\textcolor{purple}{\textbf{\textit{X}}}}}^{\prime\prime})) \\
  &\boldsymbol{F}_{h\_{\textcolor{cyan}{\textbf{\textit{Y}}}}}^{\prime\prime}, \boldsymbol{F}_{v\_{\textcolor{cyan}{\textbf{\textit{Y}}}}}^{\prime\prime}=Sig(Conv2d(\boldsymbol{F}_{h\_{\textcolor{cyan}{\textbf{\textit{Y}}}}}^{\prime\prime})), Sig(Conv2d(\boldsymbol{F}_{v\_{\textcolor{cyan}{\textbf{\textit{Y}}}}}^{\prime\prime})) \\
  &\boldsymbol{F}_h^{\prime}=\boldsymbol{F}_h \times \boldsymbol{F}_{h\_{\textcolor{purple}{\textbf{\textit{X}}}}}^{\prime\prime} \times \boldsymbol{F}_{h\_{\textcolor{cyan}{\textbf{\textit{Y}}}}}^{\prime\prime} \quad \textcolor{lightgray}{\textit{\# Re-weight}} \\
  &\boldsymbol{F}_v^{\prime}=\boldsymbol{F}_v \times \boldsymbol{F}_{v\_{\textcolor{purple}{\textbf{\textit{X}}}}}^{\prime\prime} \times \boldsymbol{F}_{v\_{\textcolor{cyan}{\textbf{\textit{Y}}}}}^{\prime\prime}  \quad \textcolor{lightgray}{\textit{\# Re-weight}}
\end{aligned}
\end{equation}
 
In general, the average pooling operation can extract more robust feature representations. The horizontal and vertical lines in the distorted document combined with the coordinate information in the X and Y directions in the feature map help to strengthen the geometric representation of the deformed features. Experimental results (Table~\ref{tab:ablation} and~\ref{tab:abla_HV_Line}) prove the effectiveness of this fusion module.


\textbf{Loss Function}. During the model training process, we predict horizontal and vertical lines with Binary Cross-Entropy (BCE) loss \cite{Boer2005ATO}. To optimize the imbalance between the pixel ratio of the line mask and other areas, we use the line loss $\mathcal{L}_{\textit{line}}$ proposed in RDGR \cite{Jiang2022RevisitingDI} as the loss for predicting horizontal and vertical lines ($\mathcal{L}_{\textit{line}}^{h}$, $\mathcal{L}_{\textit{line}}^{v}$). This loss uses the $L_2$ loss weighted pixel proportions. In order to weight the output of different layers of the UNet decoder, we use different loss weights for each decoder layer. The losses are calculated as follows:
\begin{equation}
\begin{aligned}
  &\mathcal{L}_{\textit{line}}=\sum_{i=1}^{\textit{L}}[\mathcal{L}_{\textit{line}}^{h} + \mathcal{L}_{\textit{line}}^{v} + \frac{1}{2{L}-i}(\mathcal{L}_{\textit{bce}}^{h} + \mathcal{L}_{\textit{bce}}^{v})] \\
\end{aligned}
\end{equation}

\noindent where $L=4$ is the number of layers of the decoder. $i$ is the $i$-$th$ layer of the decoder, $i \in \{0, 1, 2, 3\}$. 

To quantify the image rectification loss of our model during training, we compute the $L_1$ distance between the predicted deformation field $\boldsymbol{\hat{G}}$ of the warped image and its ground truth $\boldsymbol{G}$. The calculation is as follows:

\begin{equation}
    \mathcal{L}_\textit{rec} = \left \| \boldsymbol{G} - \hat{\boldsymbol G} \right \| _1
\end{equation}

Our model concurrently optimizes lines mask along with 2-dimensional deformation fields in an end-to-end manner. The formula for the overall training loss is presented below:

\begin{equation}
    \mathcal{L} = \alpha\mathcal{L}_{\textit{rec}} + \mathcal{L}_{\textit{line}}
\end{equation}

\noindent where $\alpha$ represents a hyper-parameter weight, which is assigned a value of 5 to ensure an balance among various loss functions.

\begin{table*}[t]
\centering
\caption{Comparison between Doc3D\cite{Das2019DewarpNetSD} and our proposed DocDewarpHV datasets of distorted document images. Where H-Line and V-Line represent horizontal and vertical line annotations, respectively. Compared to the global-only annotations in Doc3D, our dataset provides fine-grained local annotations in both horizontal and vertical dimensions, which enables researchers to develop rectification methods that target complex local deformations.}
\label{Data_statistics}
\begin{tabular}{lccccccccc}
\toprule
\multirow{2}{*}{Dataset} & \multirow{2}{*}{Image} &\multirow{2}{*}{Resolution} & \multicolumn{2}{c}{Global Info.} & \multicolumn{2}{c}{Local Info.} &\multicolumn{2}{c}{Language} \\

& & & 3D Coord. & UV Map & H-Line & V-Line & English & Chinese \\
\midrule

Doc3D \cite{Das2019DewarpNetSD} & 102,027 & $448\times448$ & \ding{51} & \ding{51} & \ding{55} & \ding{55} & \ding{51} & \ding{55} \\
DocDewarpHV (Ours) & 114,385 & $512\times512$ & \ding{51} & \ding{51} & \ding{51} & \ding{51} & \ding{51} & \ding{51} \\
\bottomrule
\end{tabular}
\end{table*}

\begin{table*}[t]
\centering
  \caption{Comparisons on DocUNet benchmark \cite{Ma2018DocUNetDI}. ``$\uparrow$'' indicates the higher the better and ``$\downarrow$'' denotes the opposite. The best performing result is shown in \textbf{Bold} font, and the second best result is shown with an \underline{underline}. ``*'' indicates the results reproduced using the authors' publicly released code on our proposed training set (DocDewarpHV) for fair comparison.}
  \label{tab:eval_docunet}
  \begin{tabular}{rrcccccc}
    \toprule
    Method & Venue & MS-SSIM $\uparrow$ & LD $\downarrow$ & AD $\downarrow$ & ED $\downarrow$ & CER $\downarrow$ & \#Param. \\
    \midrule
    Distorted & - & 0.25 & 20.51 & 1.012 & 2111.56/1552.22 & 0.5352/0.5089 & - \\
    DewarpNet \cite{Das2019DewarpNetSD} & ICCV’19 & 0.47 & 8.39 & 0.426 & 885.90/525.45 & 0.2373/0.2102 & 86.9M \\
    RDGR \cite{Jiang2022RevisitingDI} & CVPR’22 & 0.50 & 8.51 & 0.461 & 729.52/420.25 & 0.1717/0.1559 & 51.8M \\
    DocGeoNet \cite{Feng2022GeometricRL} & ECCV’22 & 0.50 & 7.71 & 0.380 & 713.94/379.00 & 0.1821/0.1509 & 24.8M \\
    PaperEdge \cite{Ma2022LearningFD} & SIGGRAPH’22 & 0.47 & 7.99 & 0.392 & 777.76/375.60 & 0.2014/0.1541 & 36.6M \\
    FTDR \cite{Li2023ForegroundAT} & ICCV'23 & 0.50 & 8.43 & 0.376 & 697.52/450.92 & 0.1705/0.1679 & 63.6M \\
    LA-DocFlatten \cite{Li2023LayoutawareSD} & TOG'23 & \underline{0.53} & \underline{6.72} & 0.300 & 695.00/391.90 & 0.1750/0.1530  & - \\
    UVDoc \cite{UVDoc} & SIGGRAPH'23 & \textbf{0.55} & 6.79 & 0.310 & 797.92/493.13 & 0.1975/0.1611 & - \\
    DocRes \cite{Zhang2024DocResAG} & CVPR'24 & 0.47 & 9.37 & 0.471 & 912.50/500.27 & 0.2406/0.1751 & \underline{14.5M} \\
    DocTLNet \cite{Kumari2024AmIR} & IJDAR'24 & 0.51 & \textbf{6.70} & - & -/377.12 & -/0.15041 & - \\
    DocReal \cite{Yu_2024_WACV} & WACV'24 &0.50 & 7.03 & \textbf{0.286} & 730.96/\underline{360.32} & 	0.1909/\underline{0.1443} \\
    DocScanner \cite{feng2025docscanner} & IJCV'25 & 0.52 & 7.45 & 0.334 & \textbf{632.30}/390.40 & \underline{0.1650}/0.1490 & \textbf{8.5M} \\
    $\text{DocScanner}^{*}$ & IJCV'25 & 0.47 & 7.85 & 0.320 & 676.22/391.98 & 0.1740/0.1433 & \textbf{8.5M} \\
    DvD \cite{zhang2025dvd}	& SIGGRAPH’25 & \textbf{0.55} & 6.84 & \underline{0.298} & 748.04/507.2 & 0.1908/0.1789 & 151.3M \\
    D2Dewarp (Ours) & - & 0.50	& 7.71 & 0.349 & \underline{656.30}/\textbf{351.08} & \textbf{0.1543}/\textbf{0.1338} & 71.2M \\
    \bottomrule
  \end{tabular}
\end{table*}

\section{Experiments}

\subsection{Datasets}
\textbf{Training Dataset.} Since the existing public training dataset Doc3D \cite{Das2019DewarpNetSD} lacks annotations for horizontal and vertical lines, and obtaining these two lines through visual image processing method will result in several errors. 
Therefore, we refer to the data synthesis methods of DewarpNet \cite{Das2019DewarpNetSD}, DocMAE \cite{Liu2023DocMAEDI}, and LA-DocFlatten \cite{Li2023LayoutawareSD} to synthesize distorted images containing Chinese and English documents as training sets using the public rendering tool blender\footnote{https://pypi.org/project/bpy/}. English documents come from the development set of PubLayNet \cite{Zhong2019PubLayNetLD}. 
Chinese documents are obtained from the CDLA \cite{li2021cdla}, CDDOD \cite{li2020cross}, and test sets of M6Doc \cite{Cheng_2023_CVPR}. 
These document datasets all have layout annotation information. We use the public OCR engine PaddleOCR\footnote{https://github.com/PaddlePaddle/PaddleOCR} to detect the coordinate of text on the original scanned images mentioned above.
By merging the boundaries of document page and layout, and information of text lines, the masks with horizontal and vertical lines can be obtained separately. 
Using the same rendering method to generate distorted results for these two masks and the scanned image. 
Furthermore, each distorted image incorporates diverse realistic textures background and natural lighting conditions to enhance its realism.

About 110K warped images are synthesized, named {\itshape DocDewarpHV}. Some samples are shown in Figure~\ref{fig:3views}. 
Table~\ref{Data_statistics} shows the statistics of our proposed warped training dataset and the comparison with Doc3D \cite{Das2019DewarpNetSD}.
During the synthesis process, we randomly add background as interference. In addition to distorted  images, it also contains 3D coordinates, UV map, horizontal and vertical line masks. In order to promote further exploration of document rectification task in the research community, we will open source this training dataset.

\textbf{DocUNet Benchmark.} The benchmark introduced by Ma \etal\cite{Ma2018DocUNetDI}, which is the first application of deep learning in this task. This benchmark has been widely used in recent research, featuring 130 distorted images of documents in natural settings, captured using mobile devices. The dataset encompasses a diverse array of document types, including bills, papers, and posters.

\textbf{DIR300 Benchmark.}
The dataset assembled and released by Feng \etal\cite{Feng2022GeometricRL}, consists of 300 authentic English document samples, creating a more extensive test set compared to the DocUNet Benchmark dataset \cite{Ma2018DocUNetDI}. Beyond merely amplifying the degree of document distortion, the dataset encompasses more complex backgrounds and a range of lighting conditions. 

\textbf{DocReal Benchmark.} This benchmark \cite{Yu_2024_WACV} contains a total of 200 distorted document images. Compared with DocUNet \cite{Ma2018DocUNetDI} and DIR300 \cite{Feng2022GeometricRL} benchmarks, it is currently the largest number of distorted Chinese documents. Image types cover contracts, bills, and books. These images have complex real backgrounds and various deformations, and have the characteristics of more scenes, greater reality, and larger deformation range.

\begin{figure*}
\begin{center}
\includegraphics[width=1.0\linewidth]{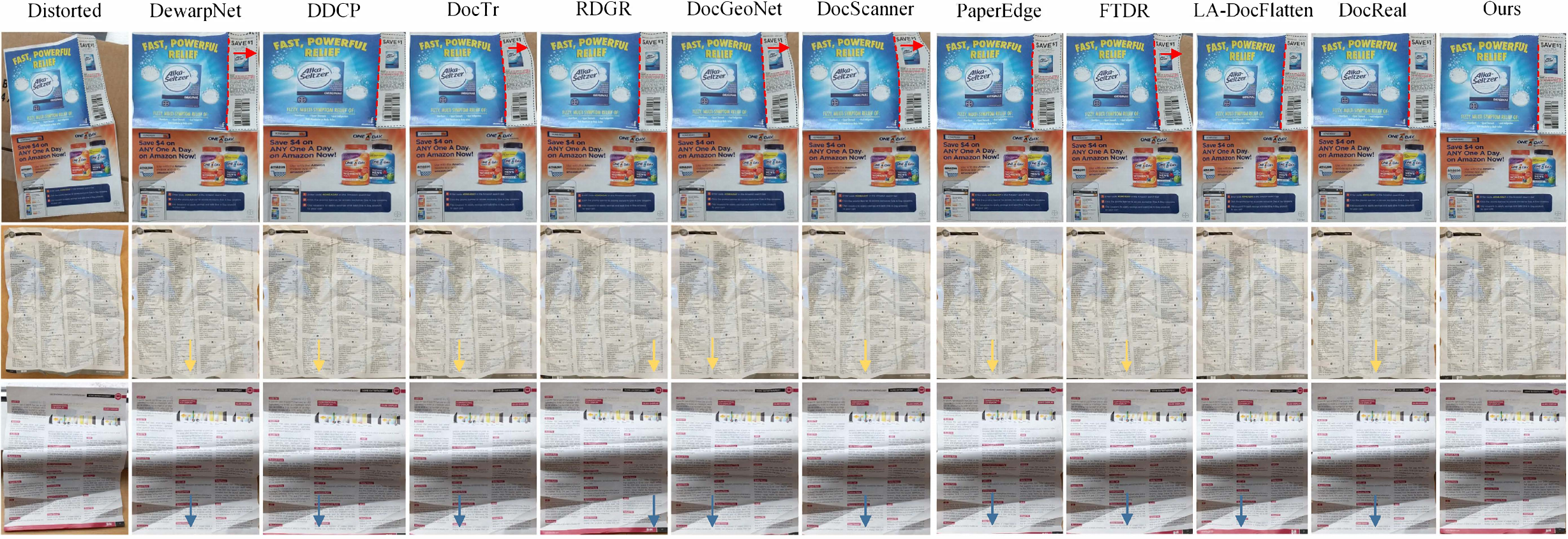}
\end{center}
\vspace{-4mm}
  \caption{Qualitative visualization comparison between our proposed and existing methods. The first column is the input distorted document image, and the last column is the dewarping result of our method. The middle columns are the effects of other previous methods. Colored arrows and dashed lines highlight differences.}
  \label{fig:DocUNet-vis}
  \vspace{-2mm}
\end{figure*}

\subsection{Evaluation metrics}
\emph{MS-SSIM, LD and AD.} Our evaluation employs metrics widely used in prior methods DocUNet, DocTr, DocGeoNet, FTDR, LA-DocFlatten, DocRes and DocScanner. Multi-Scale Structural Similarity (MS-SSIM) \cite{Wang2004ImageQA} enhances SSIM by evaluating contrast and structure across multiple scales and measuring luminance only at the final scale. Local Distortion (LD) \cite{You2016MultiviewRO} quantifies the average local deformation per pixel between the corrected image and an authentic scanned counterpart, utilizing SIFT flow \cite{Liu2011SIFTFD} for analysis. Aligned Distortion (AD) \cite{Ma2022LearningFD} optimized the problem that MS-SSIM and LD are sensitive to minor global changes.

\emph{ED and CER. }Our method's dewarping quality is assessed using OCR performance. Edit Distance (ED) is defined as the minimum number of operations needed to transform one string into another, including insertions (\textit{i}), deletions (\textit{d}), and substitutions (\textit{s}). The Character Error Rate (CER) is calculated using: ($\textit{i} + \textit{d} + \textit{s}) / \textit{M}$, where \textit{M} is the total number of characters in the reference string. Following previous methods, our evaluation on the DocUNet involves 50 and 60 images, 90 text-rich images on DIR300 and all 200 images on the DocReal benchmark. 

\subsection{Implementation details}
During training, input images are cropped and then resized to $448 \times 448$. optimization is performed using the AdamW optimizer \cite{Loshchilov2017DecoupledWD} with batch size 28. The learning rate begins at a maximum of $1\times 10^{-4}$ and decreases to a minimum of $1\times 10^{-7}$, employing a cosine learning rate decay strategy. The model undergoes training for 80 epochs with a warm-up phase of 10,000 steps. Following existing methods, evaluation metrics MS-SSIM, LD, and AD are computed using Matlab R2019a. OCR performance assessment on the two English benchmarks DocUNet and DIR300 are conducted using Tesseract v5.0.1.2022011 \cite{Smith2007AnOO} with pytesseract v0.3.8 \footnote{https://pypi.org/project/pytesseract/}. Since PaddleOCR is open source and available, and has shown excellent performance in Chinese text recognition, we use it for evaluation on the Chinese benchmark DocReal \cite{Yu_2024_WACV}. %


\subsection{Experimental results}
Since Doc3D \cite{Das2019DewarpNetSD} lacks fine-grained line annotations, in order to ensure a fair comparison and exclude performance gains from the dataset itself, we retrained the latest method DocScanner \cite{feng2025docscanner} on our proposed DocDewarpHV dataset, marked with ``*'' in the tables.

\begin{figure}[t]
\begin{center}
\includegraphics[width=0.9\linewidth]{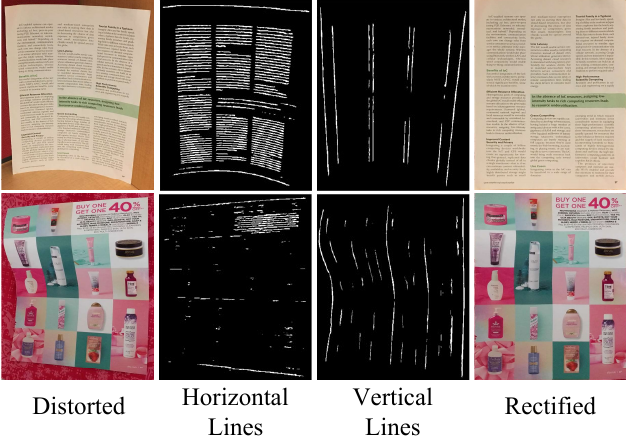}
\end{center}
  \vspace{-3mm}
  \caption{\textbf{Predict Results} Visualization. The first row corresponds to the \emph{text-dense} document, and the second row to the \emph{text-sparse} correction results.}
  \label{fig:predict_vis}
\end{figure}

\textbf{Results on DocUNet Benchmark}. We show the comparison between our model and previous rectification methods in Table~\ref{tab:eval_docunet}. Our network almost completely surpasses previous methods in terms of CER and ED. Compared with the \emph{textline-introducing methods} RDGR \cite{Jiang2022RevisitingDI}, DocGeoNet \cite{Feng2022GeometricRL}, and FTDR \cite{Li2023ForegroundAT}, our method improves the CER by at least 9.5\% and 11.3\% on 50 and 60 images respectively. Compared with the \emph{layout-focused method} LA-DocFlatten \cite{Li2023LayoutawareSD}, we outperform by 10.4\% on ED and 12.5\% on CER for the 60-image test setting.
Our method shows marginal advantages on MS-SSIM, LD, and ED metrics. The main reason is that the DocUNet \cite{Ma2018DocUNetDI} benchmark contains more text-sparse (figure-heavy) samples than the other two datasets. For text-sparse images (as opposed to text-dense document images), the predicted lines appear sparsely distributed only along document boundaries and chart regions, providing weaker guidance for the rectification model. The difference in line guidance information is illustrated in the first example (text-dense) and the second example (text-sparse) in Figure~\ref{fig:predict_vis}. 

\textbf{Results on DIR300 Benchmark}. As shown in Table~\ref{tab:eval_dir300}, the best results are achieved for most indicators on DIR300. 
Compared with the LA-DocFlatten \cite{Li2023LayoutawareSD} introduced layout information, we enhance AD by 4.6\%. On OCR evaluation metrics, our method outperforms the current state-of-the-art method DocTLNet \cite{Kumari2024AmIR} by 3.2\% on CER and 5.1\% on ED.
The warped images in this dataset contain more text content. For rich text distorted images, the denser the text lines, the better the rectification performance of our method. 

\begin{table}[t]
\centering
  \caption{Comparisons rectification performance on DIR300 benchmark \cite{Feng2022GeometricRL}. ``MS'' is MS-SSIM.}
  \label{tab:eval_dir300}
  \begin{tabular}{rccc}
    \toprule
    Method & MS$\uparrow$ & LD/AD$\downarrow$ & ED/CER$\downarrow$ \\
    \midrule
    Distorted & 0.32 & 39.58/0.771 & 1500.56/0.523 \\
    DocGeoNet & 0.64 & 6.40/0.242 & 664.96/0.219 \\
    PaperEdge & 0.58 & 8.00/0.255 & 508.73/0.207 \\
    FTDR & 0.61 & 7.68/0.244 & 652.80/0.212 \\
    LA-DocFlatten & \underline{0.65} & \textbf{5.70}/\underline{0.195} & 511.13/0.189 \\
    UVDoc & 0.62 & 7.71/0.218 & 605.24/0.260 \\
    DocTLNet & \textbf{0.66} & 5.75/- & \underline{482.57}/\underline{0.178} \\
    DocScanner & 0.62 & 7.06/0.225 & 562.72/0.194 \\
    $\text{DocScanner}^{*}$ & 0.58 & 6.96/0.205 & 521.30/0.204 \\
    D2Dewarp (Ours) & \underline{0.65} & \underline{5.73}/\textbf{0.186} & \textbf{466.94}/\textbf{0.168}
 \\
    \bottomrule
  \end{tabular}
\end{table}

\begin{table}[t]
\centering
  \caption{Comparisons rectification performance on DocReal benchmark \cite{Yu_2024_WACV}. The data in this table evaluated by ourself from the released code or rectified results of previous methods.}
  \label{tab:eval_DocReal}
  \begin{tabular}{rccc}
    \toprule
    Method & MS$\uparrow$ & LD/AD$\downarrow$ & ED/CER$\downarrow$ \\
    \midrule
    Distorted & 0.32 & 35.79/0.969 & 333.52/0.4432 \\
    DocGeoNet & 0.55 & 12.24/0.314 & 216.68/0.2974 \\
    PaperEdge & 0.52 & 11.47/0.303 & \underline{186.86}/\textbf{0.2451} \\
    RDGR & 0.54 & 11.47/0.333 & 188.00/0.2768 \\
    UVDoc & 0.55 & 11.40/0.274 & 203.12/0.2688 \\
    DocReal & \underline{0.56} & \underline{9.83}/\underline{0.238} & \textbf{184.54}/\underline{0.2485} \\
    DocScanner & 0.54 & 12.36/0.308 & 196.14/0.2709 \\
    D2Dewarp (Ours) & \textbf{0.58} & \textbf{8.69}/\textbf{0.227} & 191.25/0.2588 \\
    \bottomrule
  \end{tabular}
  \vspace{-3mm}
\end{table}

\textbf{Results on DocReal Benchmark}. Table~\ref{tab:eval_DocReal} shows the quantitative evaluation results of distorted image correction on the DocReal benchmark. Our method brings significant improvements in the three metrics of MS-SSIM (3.6\%), LD (11.6\%) and AD (4.6\%). However, our method still needs to be further improved in ED and CER. By our analysis, there are always differences in line heights between English and Chinese documents, especially text lines. A viable enhancement involves substituting the AvgPool in the HV Fusion Module (Figure~\ref{fig:fusion_module}) with a parameterized learnable convolution downsampling module. This replacement preserves more line features compared to direct pooling reduction to 1-Dimension outputs along width or height direction, but at increased computational and parametric overhead.

\begin{figure}[t]
\begin{center}
\includegraphics[width=0.9\linewidth]{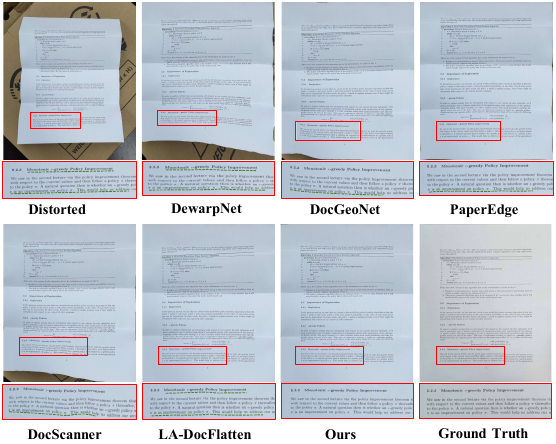}
\end{center}
  \vspace{-3mm}
  \caption{Qualitative visual comparison of local rectification in the horizontal direction between our proposed D2Dewarp and existing methods, including DewarpNet \cite{Das2019DewarpNetSD}, DocGeoNet \cite{Feng2022GeometricRL}, PaperEdge \cite{Ma2022LearningFD}, DocScanner \cite{feng2025docscanner}, and LA-DocFlatten \cite{Li2023LayoutawareSD}. The results demonstrate the superior capability of our method in correcting horizontal distortions. In the figure, ``Distorted'' represents the input warped image, while ``Ground Truth'' corresponds to the flattened reference. Red boxes mark local regions and green dashed lines show differences.}
  \vspace{-4mm}
  \label{fig:DIR300_vis_line1}
\end{figure}

The qualitative results are shown in Figure~\ref{fig:DocUNet-vis}. Colored arrows and dashed lines highlight rectification differences across comparative methods. The examples in the figure show that although our method only segments horizontal and vertical lines, it can also effectively extract foreground boundaries.

To better demonstrate the superior rectification performance of our method in the horizontal direction compared to existing methods DewarpNet \cite{Das2019DewarpNetSD}, DocGeoNet \cite{Feng2022GeometricRL}, PaperEdge \cite{Ma2022LearningFD}, DocScanner \cite{feng2025docscanner}, and LA-DocFlatten \cite{Li2023LayoutawareSD}, we cropped and enlarged the local results from text-dense, warped document images, as shown in the red rectangles in Figure ~\ref{fig:DIR300_vis_line1}. The results show that our method produces visibly straighter text lines with less curvature. This visual evidence further validates the effectiveness of our proposed horizontal constraints.
To more precisely examine the line results predicted and the rectified output by our model, we visualize these two lines as depicted in Figure ~\ref{fig:predict_vis}. For \emph{text-sparse} distorted documents, our method effectively distinguishes document foreground from background using detected document boundaries (bottom row in Figure~\ref{fig:predict_vis}).

\textbf{Speed.} Our model processes an image in 0.39 seconds on average, placing its speed between FTDR (0.6s) and DocGeoNet (0.3s). It is significantly faster than RDGR (1.87s) but slower than DocScanner (0.02s). The additional overhead stems from segmenting optimal dual lines and using an attention-based fusion module, which are essential for performance. This design balances correction quality against processing speed.

\begin{figure}[t]
\begin{center}
\includegraphics[width=0.9\linewidth]{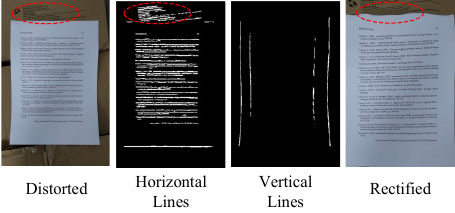}
\end{center}
  \vspace{-2mm}
  \caption{\textbf{Bad Case}. This case shows background text lines inducing erroneous horizontal line segmentation.}
    \vspace{-4mm}
  \label{fig:bad_case}
\end{figure}

\begin{table}[t]
\centering
  \caption{Ablation studies of HV Fusion Module on DocUNet, DIR300 and DocReal benchmarks. HV is HV Fusion Module. ``\Checkmark'' indicates that the feature fusion module of horizontal or vertical lines is used. ``\ding{55}'' is the opposite.}
  \label{tab:ablation}
  \begin{tabular}{lccccc}
    \toprule
    Bench. & HV & MS$\uparrow$ & LD/AD$\downarrow$ & ED/CER$\downarrow$ \\
    \midrule
    \multirow{2}{*}{DocUNet \cite{Ma2018DocUNetDI}} & \ding{55} & 0.50 & \textbf{7.6}/\textbf{0.343} & 401.1/0.1439 \\
    & \Checkmark & \textbf{0.50} & 7.7/0.349 & \textbf{351.1}/\textbf{0.1338} \\
    \midrule
    \multirow{2}{*}{DIR300 \cite{Feng2022GeometricRL}} & \ding{55} & 0.64 & \textbf{5.6}/0.193 & 509.8/0.1682 \\
    & \Checkmark & \textbf{0.65} & 5.7/\textbf{0.186} & \textbf{466.9}/\textbf{0.1676} \\
    \midrule
    \multirow{2}{*}{DocReal \cite{Yu_2024_WACV}} & \ding{55} & 0.56 & 9.2/0.252 & 199.7/0.2665 \\
    & \Checkmark & \textbf{0.58} & \textbf{8.7}/\textbf{0.227} & \textbf{191.3}/\textbf{0.2588} \\
    \bottomrule
  \end{tabular}
\end{table} 

\begin{table}[t]
\centering
  \caption{Ablation studies of horizontal or vertical line (H-Line, V-Line) predicted or not on DocReal Benchmark.}

  \label{tab:abla_HV_Line}
  \begin{tabular}{cccccc}
    \toprule
    H-Line & V-Line & MS$\uparrow$ & LD/AD$\downarrow$ & ED/CER$\downarrow$ \\
    \midrule
    \Checkmark & \ding{55} & 0.58 &	9.01/0.253 & 197.98/0.267
  \\
    \ding{55} & \Checkmark & 0.57 &	9.00/0.243 & 201.12/0.265
 \\
    \Checkmark & \Checkmark & \textbf{0.58} & \textbf{8.69}/\textbf{0.227} & \textbf{191.25}/\textbf{0.259} \\
    \bottomrule
  \end{tabular}
  \vspace{-4mm}
\end{table}

\textbf{Ablation Studies}. \emph{For HV Fusion Module}, 
to validate D2Dewarp efficacy, we perform ablation studies assessing the HV Fusion Module's impact. The HV Module was replaced with the same number of self-attention layers for a fair comparison. As quantified in Table~\ref{tab:ablation}, this module significantly enhances dewarping performance, particularly improving document readability metrics (ED and CER). While LD and AD show slight fluctuations, these results indicate the module prioritizes readability optimization.

\emph{For Dual Line Features}, 
we further demonstrate the differential impact of horizontal (H-Line) and vertical (V-Line) features through controlled ablation experiments (Table~\ref{tab:abla_HV_Line}). For fair comparison, the same number of self-attention layers is used in the ablation experiments. The results show that using both H-Line and V-Line outperforms using only single-dimensional line. This demonstrates that the two dimensions of information can effectively complement and constrain each other, optimizing deformation perception from different perspectives. 

\textbf{Bad Case Analysis and Limitations}. 
Our approach achieves substantial gains but retains limitations. Figure~\ref{fig:bad_case} shows a representative failure case where interfering background text lines (red dashed ellipse) trigger erroneous horizontal line detections. The text lines in the background prevent accurate document foreground isolation, leading to rectified images retaining extraneous background elements. This limitation can be alleviated by incorporating global features such as foreground or UV maps in the future.

\section{Conclusion}
We introduce the \emph{DocDewarpHV} dataset, featuring warped document images with fine-grained horizontal and vertical annotations. Our \emph{D2Dewarp} approach perceives bidirectional line features to automatically learn distortion patterns and enable feature complementarity. The Horizontal-Vertical Fusion Module enhances cross-directional feature interaction, improving geometric deformation representation. Quantitative and ablation studies validate the model's efficacy. By jointly modeling horizontal and vertical lines, our method captures intrinsic directional relationships, enhancing document flatness and readability. 



{
    \small
    \bibliographystyle{ieeenat_fullname}
    \bibliography{main}

@String(CVPR= {IEEE Conf. Comput. Vis. Pattern Recog.})

@String(ICCV= {Int. Conf. Comput. Vis.})

@String(ICME = {Int. Conf. Multimedia and Expo})

@String(CVPR  = {CVPR})

@String(ICCV  = {ICCV})

@String(ICME  =	{ICME})

@article{Wada1997ShapeFS,
  title={Shape from Shading with Interreflections Under a Proximal Light Source: Distortion-Free Copying of an Unfolded Book},
  author={Toshikazu Wada and Hiroyuki Ukida and Takashi Matsuyama},
  journal={International Journal of Computer Vision},
  year={1997},
  volume={24},
  pages={125-135},
  url={https://api.semanticscholar.org/CorpusID:9368922}
}

@article{Courteille2007ShapeFS,
  title={Shape from shading for the digitization of curved documents},
  author={Fr{\'e}d{\'e}ric Courteille and Alain Crouzil and Jean-Denis Durou and Pierre Gurdjos},
  journal={Machine Vision and Applications},
  year={2007},
  volume={18},
  pages={301-316},
  url={https://api.semanticscholar.org/CorpusID:1344849}
}

@article{Liang2008GeometricRO,
  title={Geometric Rectification of Camera-Captured Document Images},
  author={Jian Liang and Daniel DeMenthon and David S. Doermann},
  journal={IEEE Transactions on Pattern Analysis and Machine Intelligence},
  year={2008},
  volume={30},
  pages={591-605},
  url={https://api.semanticscholar.org/CorpusID:1599704}
}

@article{Brown2006GeometricAS,
  title={Geometric and shading correction for images of printed materials using boundary},
  author={M. S. Brown and Yau-Chat Tsoi},
  journal={IEEE Transactions on Image Processing},
  year={2006},
  volume={15},
  pages={1544-1554},
  url={https://api.semanticscholar.org/CorpusID:7378147}
}

@article{Tsoi2007MultiViewDR,
  title={Multi-View Document Rectification using Boundary},
  author={Yau-Chat Tsoi and M. S. Brown},
  journal={2007 IEEE Conference on Computer Vision and Pattern Recognition},
  year={2007},
  pages={1-8},
  url={https://api.semanticscholar.org/CorpusID:17470306}
}

@inproceedings{Meng2018ExploitingVF,
  title={Exploiting Vector Fields for Geometric Rectification of Distorted Document Images},
  author={Gaofeng Meng and Yuanqi Su and Ying Wu and Shiming Xiang and Chunhong Pan},
  booktitle={European Conference on Computer Vision},
  year={2018},
  url={https://api.semanticscholar.org/CorpusID:52955312}
}

@article{Huang2015TextLE,
  title={Text line extraction of curved document images using hybrid metric},
  author={Zuming Huang and Jie Gu and Gaofeng Meng and Chunhong Pan},
  journal={2015 3rd IAPR Asian Conference on Pattern Recognition (ACPR)},
  year={2015},
  pages={251-255},
  url={https://api.semanticscholar.org/CorpusID:34690763}
}

@article{Ma2018DocUNetDI,
  title={DocUNet: Document Image Unwarping via a Stacked U-Net},
  author={Ke Ma and Zhixin Shu and Xue Bai and Jue Wang and Dimitris Samaras},
  journal={2018 IEEE/CVF Conference on Computer Vision and Pattern Recognition},
  year={2018},
  pages={4700-4709}
}

@article{Das2019DewarpNetSD,
  title={DewarpNet: Single-Image Document Unwarping With Stacked 3D and 2D Regression Networks},
  author={Sagnik Das and Ke Ma and Zhixin Shu and Dimitris Samaras and Roy Shilkrot},
  journal={2019 IEEE/CVF International Conference on Computer Vision (ICCV)},
  year={2019},
  pages={131-140}
}

@article{Liu2020GeometricRO,
  title={Geometric rectification of document images using adversarial gated unwarping network},
  author={Xiyan Liu and Gaofeng Meng and Bin Fan and Shiming Xiang and Chunhong Pan},
  journal={Pattern Recognit.},
  year={2020},
  volume={108},
  pages={107576},
  url={https://api.semanticscholar.org/CorpusID:221766100}
}

@article{Xie2020DewarpingDI,
  title={Dewarping Document Image by Displacement Flow Estimation with Fully Convolutional Network},
  author={Guo-Wang Xie and Fei Yin and Xu-Yao Zhang and Cheng-Lin Liu},
  journal={ArXiv},
  year={2020},
  volume={abs/2104.06815},
  url={https://api.semanticscholar.org/CorpusID:225444950}
}

@article{Feng2021DocTrDI,
  title={DocTr: Document Image Transformer for Geometric Unwarping and Illumination Correction},
  author={Hao Feng and Yuechen Wang and Wen-gang Zhou and Jiajun Deng and Houqiang Li},
  journal={Proceedings of the 29th ACM International Conference on Multimedia},
  year={2021}
}

@inproceedings{Xie2022DocumentDW,
  title={Document Dewarping with Control Points},
  author={Guo-Wang Xie and Fei Yin and Xu-Yao Zhang and Cheng-Lin Liu},
  booktitle={IEEE International Conference on Document Analysis and Recognition},
  year={2022}
}

@InProceedings{Yu_2024_WACV,
    author    = {Yu, Fangchen and Xie, Yina and Wu, Lei and Wen, Yafei and Wang, Guozhi and Ren, Shuai and Chen, Xiaoxin and Mao, Jianfeng and Li, Wenye},
    title     = {DocReal: Robust Document Dewarping of Real-Life Images via Attention-Enhanced Control Point Prediction},
    booktitle = {Proceedings of the IEEE/CVF Winter Conference on Applications of Computer Vision (WACV)},
    month     = {January},
    year      = {2024},
    pages     = {665-674}
}

@article{Jiang2022RevisitingDI,
  title={Revisiting Document Image Dewarping by Grid Regularization},
  author={Xiangwei Jiang and Rujiao Long and Nan Xue and Zhibo Yang and Cong Yao and Guisong Xia},
  journal={2022 IEEE/CVF Conference on Computer Vision and Pattern Recognition (CVPR)},
  year={2022},
  pages={4533-4542}
}

@article{Feng2022GeometricRL,
  title={Geometric Representation Learning for Document Image Rectification},
  author={Hao Feng and Wen-gang Zhou and Jiajun Deng and Yuechen Wang and Houqiang Li},
  journal={ArXiv},
  year={2022},
  volume={abs/2210.08161}
}

@article{Li2023ForegroundAT,
  title={Foreground and Text-lines Aware Document Image Rectification},
  author={Heng Li and Xiangping Wu and Qingcai Chen and Qianjin Xiang},
  journal={2023 IEEE/CVF International Conference on Computer Vision (ICCV)},
  year={2023},
  pages={19517-19526},
  url={https://api.semanticscholar.org/CorpusID:267026179}
}

@article{Li2023LayoutawareSD,
  title={Layout-aware Single-image Document Flattening},
  author={Pu Li and Weize Quan and Jianwei Guo and Dongming Yan},
  journal={ACM Transactions on Graphics},
  year={2023},
  volume={43},
  pages={1 - 17},
  url={https://api.semanticscholar.org/CorpusID:264039229}
}

@article{Zhang2022MariorMR,
  title={Marior: Margin Removal and Iterative Content Rectification for Document Dewarping in the Wild},
  author={Jiaxin Zhang and Canjie Luo and Lianwen Jin and Fengjun Guo and Kai Ding},
  journal={Proceedings of the 30th ACM International Conference on Multimedia},
  year={2022},
  url={https://api.semanticscholar.org/CorpusID:251040732}
}

@article{Ma2022LearningFD,
  title={Learning From Documents in the Wild to Improve Document Unwarping},
  author={Ke Ma and Sagnik Das and Zhixin Shu and Dimitris Samaras},
  journal={ACM SIGGRAPH 2022 Conference Proceedings},
  year={2022},
  url={https://api.semanticscholar.org/CorpusID:250702209}
}

@article{Das2021EndtoendPU,
  title={End-to-end Piece-wise Unwarping of Document Images},
  author={Sagnik Das and Kunwar Yashraj Singh and Jon Wu and Erhan Bas and Vijay Mahadevan and Rahul Bhotika and Dimitris Samaras},
  journal={2021 IEEE/CVF International Conference on Computer Vision (ICCV)},
  year={2021},
  pages={4248-4257},
  url={https://api.semanticscholar.org/CorpusID:243947553}
}

@article{feng2023doctrp,
  title={Deep Unrestricted Document Image Rectification},
  author={Feng, Hao and Liu, Shaokai and Deng, Jiajun and Zhou, Wengang and Li, Houqiang},
  journal={IEEE Transactions on Multimedia},
  year={2023}
}

@article{Liu2023DocMAEDI,
  title={DocMAE: Document Image Rectification via Self-supervised Representation Learning},
  author={Shaokai Liu and Hao Feng and Wen-gang Zhou and Houqiang Li and Cong Liu and Feng Wu},
  journal={2023 IEEE International Conference on Multimedia and Expo (ICME)},
  year={2023},
  pages={1613-1618},
  url={https://api.semanticscholar.org/CorpusID:258236418}
}

@article{Ioffe2015BatchNA,
  title={Batch Normalization: Accelerating Deep Network Training by Reducing Internal Covariate Shift},
  author={Sergey Ioffe and Christian Szegedy},
  journal={ArXiv},
  year={2015},
  volume={abs/1502.03167},
  url={https://api.semanticscholar.org/CorpusID:5808102}
}

@inproceedings{Nair2010RectifiedLU,
  title={Rectified Linear Units Improve Restricted Boltzmann Machines},
  author={Vinod Nair and Geoffrey E. Hinton},
  booktitle={International Conference on Machine Learning},
  year={2010},
  url={https://api.semanticscholar.org/CorpusID:15539264}
}

@article{Boer2005ATO,
  title={A Tutorial on the Cross-Entropy Method},
  author={P. T. de Boer and Dirk P. Kroese and Shie Mannor and Reuven Y. Rubinstein},
  journal={Annals of Operations Research},
  year={2005},
  volume={134},
  pages={19-67},
  url={https://api.semanticscholar.org/CorpusID:110510}
}

@article{Liu2011SIFTFD,
  title={SIFT Flow: Dense Correspondence across Scenes and Its Applications},
  author={Ce Liu and Jenny Yuen and Antonio Torralba},
  journal={IEEE Transactions on Pattern Analysis and Machine Intelligence},
  year={2011},
  volume={33},
  pages={978-994},
  url={https://api.semanticscholar.org/CorpusID:10458500}
}

@article{Wang2004ImageQA,
  title={Image quality assessment: from error visibility to structural similarity},
  author={Zhou Wang and Alan Conrad Bovik and Hamid R. Sheikh and Eero P. Simoncelli},
  journal={IEEE Transactions on Image Processing},
  year={2004},
  volume={13},
  pages={600-612},
  url={https://api.semanticscholar.org/CorpusID:207761262}
}

@article{You2016MultiviewRO,
  title={Multiview Rectification of Folded Documents},
  author={Shaodi You and Yasuyuki Matsushita and Sudipta Sinha and Yu-Ki Bou and Katsushi Ikeuchi},
  journal={IEEE Transactions on Pattern Analysis and Machine Intelligence},
  year={2016},
  volume={40},
  pages={505-511},
  url={https://api.semanticscholar.org/CorpusID:9905889}
}

@inproceedings{Loshchilov2017DecoupledWD,
  title={Decoupled Weight Decay Regularization},
  author={Ilya Loshchilov and Frank Hutter},
  booktitle={International Conference on Learning Representations},
  year={2017},
  url={https://api.semanticscholar.org/CorpusID:53592270}
}

@article{Smith2007AnOO,
  title={An Overview of the Tesseract OCR Engine},
  author={Raymond W. Smith},
  journal={Ninth International Conference on Document Analysis and Recognition (ICDAR 2007)},
  year={2007},
  volume={2},
  pages={629-633},
  url={https://api.semanticscholar.org/CorpusID:7038773}
}

@InProceedings{Hou_2021_CVPR,
    author    = {Hou, Qibin and Zhou, Daquan and Feng, Jiashi},
    title     = {Coordinate Attention for Efficient Mobile Network Design},
    booktitle = {Proceedings of the IEEE/CVF Conference on Computer Vision and Pattern Recognition (CVPR)},
    month     = {June},
    year      = {2021},
    pages     = {13713-13722}
}

@inproceedings{ronneberger2015u,
  title={U-net: Convolutional networks for biomedical image segmentation},
  author={Ronneberger, Olaf and Fischer, Philipp and Brox, Thomas},
  booktitle={Medical image computing and computer-assisted intervention--MICCAI 2015: 18th international conference, Munich, Germany, October 5-9, 2015, proceedings, part III 18},
  pages={234--241},
  year={2015},
  organization={Springer}
}

@article{Zhong2019PubLayNetLD,
  title={PubLayNet: Largest Dataset Ever for Document Layout Analysis},
  author={Xu Zhong and Jianbin Tang and Antonio Jimeno-Yepes},
  journal={2019 International Conference on Document Analysis and Recognition (ICDAR)},
  year={2019},
  pages={1015-1022},
  url={https://api.semanticscholar.org/CorpusID:201124789}
}

@InProceedings{Cheng_2023_CVPR,
    author    = {Cheng, Hiuyi and Zhang, Peirong and Wu, Sihang and Zhang, Jiaxin and Zhu, Qiyuan and Xie, Zecheng and Li, Jing and Ding, Kai and Jin, Lianwen},
    title     = {M6Doc: A Large-Scale Multi-Format, Multi-Type, Multi-Layout, Multi-Language, Multi-Annotation Category Dataset for Modern Document Layout Analysis},
    booktitle = {Proceedings of the IEEE/CVF Conference on Computer Vision and Pattern Recognition (CVPR)},
    month     = {June},
    year      = {2023},
    pages     = {15138-15147}
}

@inproceedings{li2020cross,
  title={Cross-Domain Document Object Detection: Benchmark Suite and Method},
  author={Li, Kai and Wigington, Curtis and Tensmeyer, Chris and Zhao, Handong and Barmpalios, Nikolaos and Morariu, Vlad I and Manjunatha, Varun and Sun, Tong and Fu, Yun},
  booktitle={Proceedings of the IEEE/CVF Conference on Computer Vision and Pattern Recognition},
  pages={12915--12924},
  year={2020}
}

@misc{li2021cdla,
  title={Cdla: A chinese document layout analysis (cdla) dataset},
  author={Li, Hang},
  year={2021}
}

@ARTICLE{10327775,
  author={Liu, Shaokai and Feng, Hao and Zhou, Wengang},
  journal={IEEE Transactions on Circuits and Systems for Video Technology}, 
  title={Rethinking Supervision in Document Unwarping: A Self-Consistent Flow-Free Approach}, 
  year={2024},
  volume={34},
  number={6},
  pages={4817-4828},
  doi={10.1109/TCSVT.2023.3336068}}

@inproceedings{UVDoc,
title={{UVDoc}: Neural Grid-based Document Unwarping},
author={Floor Verhoeven and Tanguy Magne and Olga Sorkine-Hornung},
booktitle = {SIGGRAPH ASIA, Technical Papers},
year = {2023},
url={https://doi.org/10.1145/3610548.3618174}
}

@article{Zhang2024DocResAG,
  title={DocRes: A Generalist Model Toward Unifying Document Image Restoration Tasks},
  author={Jiaxin Zhang and Dezhi Peng and Chongyu Liu and Peirong Zhang and Lianwen Jin},
  journal={2024 IEEE/CVF Conference on Computer Vision and Pattern Recognition (CVPR)},
  year={2024},
  pages={15654-15664},
  url={https://api.semanticscholar.org/CorpusID:269614217}
}

@article{Kumari2024AmIR,
  title={Am I readable? Transfer learning based document image rectification},
  author={Pooja Kumari and Sukhendu Das},
  journal={Int. J. Document Anal. Recognit.},
  year={2024},
  volume={27},
  pages={433-446},
  url={https://api.semanticscholar.org/CorpusID:269997375}
}

@article{Wei2024GeneralOT,
  title={General OCR Theory: Towards OCR-2.0 via a Unified End-to-end Model},
  author={Haoran Wei and Chenglong Liu and Jinyue Chen and Jia Wang and Lingyu Kong and Yanming Xu and Zheng Ge and Liang Zhao and Jian‐Yuan Sun and Yuang Peng and Chunrui Han and Xiangyu Zhang},
  journal={ArXiv},
  year={2024},
  volume={abs/2409.01704},
  url={https://api.semanticscholar.org/CorpusID:272366864}
}

@article{Liao2024DocLayLLMAE,
  title={DocLayLLM: An Efficient and Effective Multi-modal Extension of Large Language Models for Text-rich Document Understanding},
  author={Wenhui Liao and Jiapeng Wang and Hongliang Li and Chengyu Wang and Jun Huang and Lianwen Jin},
  journal={ArXiv},
  year={2024},
  volume={abs/2408.15045},
  url={https://api.semanticscholar.org/CorpusID:271963079}
}

@article{Zhang2024DocKylinAL,
  title={DocKylin: A Large Multimodal Model for Visual Document Understanding with Efficient Visual Slimming},
  author={Jiaxin Zhang and Wentao Yang and Songxuan Lai and Zecheng Xie and Lianwen Jin},
  journal={ArXiv},
  year={2024},
  volume={abs/2406.19101},
  url={https://api.semanticscholar.org/CorpusID:270764705}
}

@article{Feng2023DocPediaUT,
  title={DocPedia: Unleashing the Power of Large Multimodal Model in the Frequency Domain for Versatile Document Understanding},
  author={Hao Feng and Qi Liu and Hao Liu and Wen-gang Zhou and Houqiang Li and Can Huang},
  journal={ArXiv},
  year={2023},
  volume={abs/2311.11810},
  url={https://api.semanticscholar.org/CorpusID:265295560}
}

@article{Dai2023MataDocMA,
  title={MataDoc: Margin and Text Aware Document Dewarping for Arbitrary Boundary},
  author={Beiya Dai and Xingbiao Li and Qunyi Xie and Yulin Li and Xiameng Qin and Chengquan Zhang and Kun Yao and Junyu Han},
  journal={ArXiv},
  year={2023},
  volume={abs/2307.12571},
  url={https://api.semanticscholar.org/CorpusID:260125937}
}

@article{Fang2020ACS,
  title={A Camera Shooting Resilient Watermarking Scheme for Underpainting Documents},
  author={Han Fang and Weiming Zhang and Zehua Ma and Hang Zhou and Shan Sun and Hao Cui and Nenghai Yu},
  journal={IEEE Transactions on Circuits and Systems for Video Technology},
  year={2020},
  volume={30},
  pages={4075-4089},
  url={https://api.semanticscholar.org/CorpusID:209093479}
}

@article{Li2023CharacterAwareSA,
  title={Character-Aware Sampling and Rectification for Scene Text Recognition},
  author={Ming Li and Bin Fu and Zhengfu Zhang and Yu Qiao},
  journal={IEEE Transactions on Multimedia},
  year={2023},
  volume={25},
  pages={649-661},
  url={https://api.semanticscholar.org/CorpusID:244523246}
}

@article{Peng2022RecognitionOH,
  title={Recognition of Handwritten Chinese Text by Segmentation: A Segment-Annotation-Free Approach},
  author={Dezhi Peng and Lianwen Jin and Weihong Ma and Canyu Xie and Hesuo Zhang and Shenggao Zhu and Jing Li},
  journal={IEEE Transactions on Multimedia},
  year={2022},
  volume={25},
  pages={2368-2381},
  url={https://api.semanticscholar.org/CorpusID:246367636}
}

@article{Jin2020RUArtAN,
  title={RUArt: A Novel Text-Centered Solution for Text-Based Visual Question Answering},
  author={Zanxia Jin and Heran Wu and Chun Yang and Fang Zhou and Jingyan Qin and Lei Xiao and Xu-Cheng Yin},
  journal={IEEE Transactions on Multimedia},
  year={2020},
  volume={25},
  pages={1-12},
  url={https://api.semanticscholar.org/CorpusID:225067381}
}

@article{feng2025docscanner,
  title={DocScanner: Robust document image rectification with progressive learning},
  author={Feng, Hao and Zhou, Wengang and Deng, Jiajun and Tian, Qi and Li, Houqiang},
  journal={International Journal of Computer Vision},
  pages={1--20},
  year={2025},
  publisher={Springer}
}

@inproceedings{han2025docmamba,
  title={DocMamba: Robust Document Image Dewarping via Selective State Space Sequence Modeling},
  author={Han, Miaolin and Li, Huibin},
  booktitle={International Conference on Multimedia Modeling},
  pages={304--318},
  year={2025},
  organization={Springer}
}

@inproceedings{hertlein2025docmatcher,
  title={DocMatcher: Document Image Dewarping via Structural and Textual Line Matching},
  author={Hertlein, Felix and Naumann, Alexander and Sure-Vetter, York},
  booktitle={2025 IEEE/CVF Winter Conference on Applications of Computer Vision (WACV)},
  pages={5771--5780},
  year={2025},
  organization={IEEE}
}

@article{zhang2025enhancing,
  title={Enhancing document dewarping evaluation: A new metric with improved accuracy and efficiency},
  author={Zhang, Jiaxin and Zhang, Peirong and Peng, Dezhi and Xu, Haowei and Jin, Lianwen},
  journal={Pattern Recognition Letters},
  year={2025},
  publisher={Elsevier}
}

@inproceedings{tie2025local,
  title={Local and Global Aware Document Image Enhancement with Residual Denoising Diffusion Model},
  author={Tie, Hongrui and Li, Heng and Wu, Xiangping and Chen, Qingcai},
  booktitle={Proceedings of the 2025 International Conference on Multimedia Retrieval},
  pages={1293--1302},
  year={2025}
}

@inproceedings{yang2023docdiff,
  title={Docdiff: Document enhancement via residual diffusion models},
  author={Yang, Zongyuan and Liu, Baolin and Xxiong, Yongping and Yi, Lan and Wu, Guibin and Tang, Xiaojun and Liu, Ziqi and Zhou, Junjie and Zhang, Xing},
  booktitle={Proceedings of the 31st ACM international conference on multimedia},
  pages={2795--2806},
  year={2023}
}

@article{zhang2025dvd,
  title={DvD: Unleashing a Generative Paradigm for Document Dewarping via Coordinates-based Diffusion Model},
  author={Zhang, Weiguang and Lu, Huangcheng and Ning, Maizhen and Huang, Xiaowei and Wang, Wei and Huang, Kaizhu and Wang, Qiufeng},
  journal={arXiv preprint arXiv:2505.21975},
  year={2025}
}
}

\end{document}